# Unmasking the Reality of PII Masking Models: Performance Gaps and the Call for Accountability

Devansh Singh
AI Tech Ethics
devanshsingh262@gmail.com

Sundaraparipurnan Narayanan
AI Tech Ethics
sundar.narayanan@aitechethics.com

## Abstract

Privacy Masking is a critical concept under data privacy involving anonymization and de-anonymization of personally identifiable information (PII) while leveraging structured or unstructured data for analysis. Privacy masking techniques rely on Named Entity Recognition (NER) approaches under NLP support in identifying and classifying named entities in each text. NER approaches, however, have several limitations including (a) content sensitivity including ambiguous, polysemic, context dependent or domain specific content, (b) phrasing variabilities including nicknames and alias, informal expressions, alternative representations, emerging expressions, evolving naming conventions and (c) formats or syntax variations, typos, misspellings. One of the major limitations to the above is sparse datasets. In addition, in the context of PII masking, due to privacy concerns, diversity of PII in regional context and complexities in identifying them, make gathering or compiling such a dataset challenging. However, there are a couple of PII datasets (namely pii_masking and Bigcode datasets) that have been widely used by researchers and the open-source community to train models on PII detection or masking. These datasets have been used to train models including Piiranha and Starpii, which have been downloaded over 300k and 580k times on HuggingFace. We examine the quality of the PII masking by these models given the limitations of the datasets and of the NER approaches. We curate a dataset of 17K unique, semi-synthetic sentences containing 16 types of PII by compiling information from across multiple jurisdictions including India, U.K and U.S. We generate sentences (using language models) containing these PII at five different NER detection feature dimensions - (1) Basic Entity Recognition, (2) Contextual Entity Disambiguation, (3) NER in Noisy & Real-World Data, (4) Evolving & Novel Entities Detection and (5) Cross-Lingual or multi-lingual NER) and 1 in adversarial context. We channel the generated text into a pipeline to test accuracy of detection by MS Presidio, Piiranha and Starpii models. We present the results and exhibit the privacy exposure caused by such model use (considering the extent of lifetime downloads of these models). We conclude by highlighting the gaps in measuring performance of the models and the need for contextual disclosure in model cards for such models. We open source the dataset and enable expansion of privacy masking approaches.

# 1 Introduction

## 1.1 Overview of Personally Identifiable Information (PII) Masking

PII masking is essential for ensuring data privacy and compliance with regulations such as the General Data Protection Regulation (GDPR) and the California Consumer Privacy Act (CCPA). These regulations mandate that organizations implement robust measures to protect personal data from unauthorized access and breaches. PII masking efforts must consider factors like data sensitivity and regulatory requirements when implementing data masking solutions. The process includes conducting thorough data discovery, implementing a layered approach to data masking, regularly updating masking rules, establishing strong access controls, monitoring data masking processes, and providing ongoing training for employees [1]. Various techniques adopted for masking, including data anonymization, pseudonymization, and encryption, which collectively aim to obscure identifiable information while maintaining the utility of the data for analytical purposes [2, 3]. PII masking enables sensitive data including credit card numbers, email ids and social security numbers to be properly managed and governed, by identifying and anonymizing the private entities (PIIs) in text. PII masking leverages approaches including Named Entity Recognition (NER), regular expressions, and rule-based logics to identify PII. In some instances, these models are trained on multi-lingual data for better identification across languages.

## 1.2 Named Entity Recognition (NER) in PII masking

Named Entities (NEs) are real-life objects that are proper names and identities of interest. This approach supports Personally Identifiable Information detection and masking, by adapting to the classification context, based on predefined PII references. NER and classification is essentially a sequence labelling task where, given a sequence of tokens, a system seeks to assign labels (NE classes) to this sequence [4]. This paper primarily speaks about the PII masking models that adopt NER-based approaches along with pattern matching. While the dataset curated or the tests undertaken are not limited to NER-based approaches, the paper does not exhaustively cover the risks contributed by regular expressions or rule-based logics in PII detection.

## 1.3 Limitations of Named Entity Recognition Approach for PII Masking

NER has limitations in accurately detecting text containing ambiguity, phrasing variability, syntax variations of content. This also applies to the detection of personally identifiable information. There are 3 main categories of limitations of NER provided below:

### 1.3.1 Content sensitivity including ambiguous, polysemic, context dependent or domain specific content:

Ambiguity in languages contributes to challenges in appropriate detection of PII. While there exist disambiguation approaches, the subject is still challenging to address. For instance, special name words (e.g., titles, organization markers) and first names need to be combined with rules for boundary detection and entity type classification to avoid disambiguation, however, such an approach does not consistently work in all instances [5, 6]. Also, a ten-digit alphanumeric number could be any PII including password, passport number, driving licence number or tax identification number, thereby making it complex for the model to detect. Further, ambiguous names (representing both a place and a person) or abbreviations or non-local dependencies also add to the complexities of NER detection methods that are available currently [7]. Sentence level structures (e.g. entities like 'Office of integrity and compliance'), document level co-references (wherein an entity is introduced in one section and is referred in another section), and document level structures (e.g. text document containing tables with entity references) also contribute to challenges in detecting NER appropriately [8]. Also, entity coreference resolutions (whether entities mentioned in a text or dialogue refer to the same real-world entity), contextual/ domain centric entities may also be challenging for NER [6].

### 1.3.2 **Phrasing variabilities including nicknames and alias, informal expressions, alternative representations, emerging expressions, evolving naming conventions:**

Bad segmentation of text (United Kingdom represented as United and Kingdom) (Stanislawek et al., 2019), linguistic phrases or dialects and alternative expressions are some of the cases where NER does not detect the entities. In addition, spelling variations, naming conventions ('husband' has evolved to become 'partner'), modulating structural representations (e.g. Neymar Jr.) contribute to additional challenges in detecting NEs [4]. Similarly, special construction within sentences (e.g. 'Philippines' which is being referred to as 'the Philippines' in the following sentence: "ASEAN groups Brunei, Indonesia, Malaysia, the Philippines, Singapore, Thailand and Vietnam." , helps its detection while other countries did not get detected) may or may not get detected in some instances depending on context [8]. Furthermore, NER detection is challenged with multilingual content wherein in some languages the script or text does not have capitalization [7]. In addition, emerging expressions 'LOL', 'IMO', 'IRL' 'MIL' in short text or informal text may also be complicated to get the context of the expression for the NER models, unless specifically trained on them [6]. Also, Transformer-based models, while effective in capturing long-range dependencies, often struggle with local feature extraction, which is critical for accurate entity recognition. For example, in the statement 'The WHO released a new report on global health', WHO refers to the World Health Organization which the model will find difficult to detect as an entity considering the local features such as capitalisation, punctuation, prefixes or suffixes and immediate sequence of words.

### 1.3.3 **Formats or syntax variations, typos, misspellings:**

Named entity recognition faces challenges with detecting the appropriate entities when the text has misspellings or typos. For example: "Pollish" instead of "Polish" or annotation errors [8] . In addition, text encoded in machines that do not have appropriate fonts or styles can also lead to detection failures or misclassification [9]. In addition, if the source text is extracted from multilingual documents or from physical / scanned documents using Optical Character Recognition methods, the challenges contributed by such a system in recognizing characters and layout can also influence the impact on NER detection [4].

Data scarcity and availability of adequately annotated datasets is one of the key reasons for the above limitations in NER. It is also considered to be costly, non-portable and domain-specific, requiring human expertise while not transferring well across domains [7]. Further, to the best of our knowledge, there is no open-source PII-masking / identification dataset that is sufficiently diverse to enable efficient detection of PII across languages and geographies among other variations. The existing datasets for PII are not tailored for global context. For instance, Unified Payment Interface, the digital peer-to-peer payment mechanism has a unique identity for each user based on their user id, phone number and bank (e.g abcd@okicici). However, these are currently detected by NER as email ids possibly because they have '@' (common regular expression in emails) as part of them. Similarly, credit card numbers are typically perceived to be 16-digit numbers in a 4-4-4-4 pattern. However, the credit card numbers range from 12 digits to 19 digits. For NER models that are trained on syntaxes (rule-based logic) of credit cards, the detection accuracy for diverse (with 12 to 19 digits) credit cards is low.

## 1.4 Research Objectives and Scope

This research aims to create a diverse PII dataset for testing prominent open-source PII masking models to evaluate them for accuracy. The research also attempts to highlight the performance gaps and transparent disclosure of such performance and limitations of these models as part of public accountability. We focused on 3 open-source PII masking models for the purpose of the research. They are Piiranha, Starpii and Microsoft Presidio.

The main contributions of this research are as follows:
(A) Evaluation of current PII-masking models using a novel Maturity-Level-inspired framework which incorporates real-world complexities.
(B) Curation of a novel semi-synthetic evaluation dataset for PII-masking models.
(C) Exhibiting the limitations or failure modes of PII-masking models (and thereby providing additional evaluation considerations) and need for transparent and detailed disclosures.

## 2 Methodology

### 2.1 Data Gathering, Model Testing and Analysis

This research examines the performance gaps of prominent open-source PII masking models, emphasizing the need for accountability in their development and deployment. Firstly, we identified two key open-source datasets for PII detection and masking: the AI4Privacy dataset [10] and the BigCode PII dataset [11]. These datasets provided a range of PII instances and associated metadata. We focused on three widely used open-source models that had used such a dataset for PII detection and masking: Piiranha [12], Starpii [13], and Microsoft Presidio [14]. Microsoft Presidio has a closed-source dataset, and we do not have access to that information. These models were chosen for their downloads (Piiranha – 0.3 mn, Bigcode – 0.58 mn. We could not get the downloads of presidio), and open-source nature.

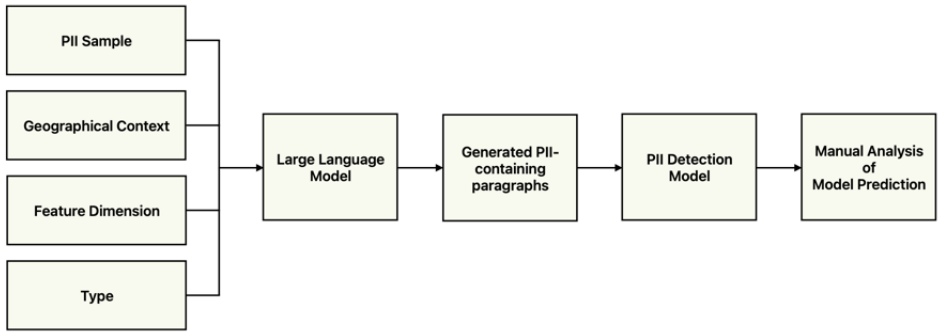

Fig. 1. Flow for dataset creation and model testing.

In order to curate a comprehensive dataset, we firstly compiled a comprehensive list of major PII identifiers and quasi-identifiers, broadly categorizing variations of these PII's into three types: (a) Syntax Variations (e.g. bank account numbers of different banks having different lengths), (b) Linguistic Variations (e.g. people's legal names appearing in book titles and song titles), and (c) Representational Variations (e.g. use of special characters like '@' instead of 'a').

In the absence of any standard framework for evaluation of PII masking models, we develop a novel framework, as shown in Fig. 1, which aims to test any PII masking model across different types of PII present in various scenarios without being restricted to a few geographies. For each type of PII, we curated variations under the relevant category (e.g. we list a few samples of different driving license numbers across geographies). We then leveraged language models to generate synthetic samples (seed PII samples) for each variation, ensuring a wide coverage of PII instances. A seed PII sample would be a single instance of a type of PII (e.g. the zip code "209010" would be a seed PII). Post which, we developed five feature dimensions and one adversarial dimension of PII detection leveraging NER approach, based on common and emerging challenges in Named Entity Recognition. These levels represented various real-life scenarios where NER may be applied for PII detection. Using the GPT-4o-mini model, we generated text paragraphs that contained each seed PII sample for the identified feature dimensions while feeding in geographical context about the type of PII (e.g. we mention that Aadhar Card number is the National Identification Number used in India). This process aimed to create diverse and challenging PII detection scenarios. Finally, we tested the text variations using the selected PII detection and masking models (Piiranha, Starpii, and Microsoft Presidio). Each model was evaluated on its ability to identify and mask PII instances accurately across the various text variations and feature dimensions. We analyzed the results to gather trends in PII identification, non-identification, and misclassification.

Although we compiled a list of variations in the PII, the curated data is semi-synthetic because additional examples for each variation and later the text paragraphs containing those samples were generated using LLMs. Table 1 describes the creation and composition of the prompts. These prompts will be available along with the dataset. However, there are limitations to this approach. In several cases, the LLM failed to generate useful outputs, or produced sentences without the intended PII.

Table 1: **LLM Prompt Composition**

| Setting | Description |
| --- | --- |
| Situation | The LLM is asked to act as a text generation bot. |
| Task | The LLM is asked to generate sentences based on the given PII, instructions and scenarios. |
| Action Instruction | The LLM is given specific sentence-generation instructions for each Feature Dimension. For example, for Contextual Entity Disambiguation, it is asked to create sentences where the PII may have multiple potential meanings. |
| Examples | Examples of expected outputs are provided. |
| Additional Guidance | Additional specific guidance on the creation of sentences is provided to enlarge the LLM's context for the task. For example, the expected output language, addition or omission of noise, etc. |
| Incentive | A reward for following the instructions and penalty for deviating is mentioned. |
| Input | The input format is defined. It contains the seed PII sample, type of PII, geographical context and information about the type of PII. |

## 2.2 **Feature Dimensions and Text Generation**

To appropriately measure the effectiveness of the PII masking model, we developed five feature dimensions and one adversarial feature dimension of PII. These dimensions were chosen based on common and emerging challenges in Named Entity Recognition and PII identification. These dimensions represent various realistic scenarios where any PII detection system should be proficient in order to be widely implemented. These dimensions are provided below in Table 2:

Table 2: **Feature Dimensions and Associated Definitions**

| Feature Dimension | Definition |
| --- | --- |
| Basic Entity Recognition | Basic Entity Recognition is a feature dimension in Named Entity Recognition where the system identifies and extracts straightforward, well-defined Personally Identifiable Information (PII) entities from text. This includes entities such as full names, dates of birth, and email addresses, which are clearly presented in the input data |
| Contextual Entity Disambiguation | Contextual Entity Disambiguation is a feature dimension in Named Entity Recognition (NER) where personally identifiable information (PII) terms can have multiple interpretations. This process requires leveraging contextual clues to accurately identify and differentiate entities based on their surrounding text |
| Noisy and Real-World Data Detection | Noisy & Real-World Data in Named Entity Recognition refers to the presence of imperfections and variations in data that reflect real-life scenarios, posing challenges to accurate PII masking using NER. This includes typographical errors, misspellings, abbreviations, and informal representations of PII, as well as PII embedded within noisy or unstructured text, such as social media posts or emails, where the context may be unclear or cluttered with irrelevant information |
| Evolving & Novel Entities Detection | Evolving & Novel Entities in Named Entity Recognition (NER) refers to the capability of identifying and processing emerging or unconventional types of Personally Identifiable Information (PII). This includes masked or obfuscated PII (e.g., partial phone numbers or anonymized names), indirect identifiers that can lead to individual identification when combined, and PII associated with new technologies or communication platforms, reflecting the dynamic nature of data privacy and identification in the digital age. |
| Cross-Lingual or Multi-Lingual NER | Cross-Lingual or Multi-Lingual Named Entity Recognition (NER) refers to the capability of a system to identify and classify named entities in text that is written in multiple languages. This level of NER involves recognizing entities such as names, organizations, locations, and other specific information across different languages, including scenarios where multiple languages are used within the same text or conversation. |

In addition, we developed an adversarial context generation with capability that intentionally confuses or misleads NER models by employing various techniques. This includes the capability to create minor perturbations and homoglyphs aim to hinder PII recognition by introducing subtle changes or using visually similar characters, and camouflaging PII and contextual manipulation involve embedding or altering the PII's context to mislead the model.

We generated text paragraphs for each sample PII leveraging language model (GPT4-o-mini) using a prompt-based approach for all the feature dimensions. Since context words are a major contributor to the performance of NER models, we generated paragraphs of two types - with and without explicit reference to PII type in order to evaluate models in both the scenarios. We referred to the text generated with explicit reference to PII type as 'Type 1' feature dimensions and the ones without explicit references as 'Type 2' feature dimensions. For example, a Type 1 text paragraph may be "Order was delivered to Zip Code 209011" while a Type 2 text paragraph may be "Order was delivered to 209011". The same approach was adopted for generating texts that have adversarial context. To ensure that the model text generation results are diverse, we included a random scenario (one of 20 scenarios including healthcare, social media, financial transactions, transportation, e-commerce, customer service). This enabled the model to generate diverse results for samples from the same variation of the PII. The generations contained the seed PII sample and additional PII automatically generated based on context provided for each variation of PII. Hence, each unit of text generated contained more than 1 PII (more than the seed PII sample). Furthermore, we added an instruction to the text generation prompt for the representational variations category requiring the model to retain the seed PII sample as a string in its generation, to avoid transformation of the content during generation. While care was taken to ensure that the text generators consistently generate the text containing the seed PII samples, there were approx. 30% of instances wherein the text generated did not contain the PII samples based on regex validation. These instances were not considered for testing.

## 2.3 Evaluation Considerations

We aim to assess the overall capability of PII masking models and their performance in diverse contexts and scenarios. We evaluated the models based on 3 key factors, namely, (a) whether the model detected any PII in the text, (b) whether the model detects the correct seed PII samples, (c) Whether the model misclassified in detecting PII (both seed PII and other generated PII in the text). Since there are no standard evaluation metrics specifically designed for PII detection models, such an analysis on a diverse dataset provides an insight into the effectiveness of such systems.

# 3 Results from model testing

We ran the list of the generated text paragraphs after eliminating aberrations (the instances where the seed PII sample was not present in them due to the LLM not following the instructions perfectly) on the three identified models to obtain predictions. The model prediction shows the substrings of the input text paragraphs which are identified as PII along with the predicted PII type. We used regex matching between the prediction and the seed PII to find out whether the model was able to detect the correct string or not. Overall, there were 17k samples (generated text paragraphs) tested across all mentioned feature dimensions. All the tested samples contained the seed PII, as validated using regex before feeding them through the models. When run through the 3 models, we obtained 51k (3 * 17k) model outputs.

## 3.1 Non-Identification of any PII in the text by the models

If the model outputs an empty list, it means that it did not detect any PII in the input text. Table 3 lists the numbers of non-identification of any PII in the input text for the models across all the feature dimensions. Overall, the models failed to recognize any PII in 28% of the 51k predictions.. The highest non-identification percentage is observed in Contextual Entity Disambiguation for both Type 1 and Type 2. This signifies that when the entity may have multiple interpretations, the models fail with a high chance of omitting any possible predictions when the confidence threshold is set to default. The Starpii model has high non-identification numbers possibly because of the fact that it is trained to detect PII embedded within code text files (like Java, Python). These results of Starpii on non-code text exhibit how context words, non-diverse datasets and out-of-distribution testing can significantly downgrade the performance of a PII detection model.

Table 3: Feature Dimension-wise trends of non-identification of any PII in the text by the models (represented as % of instances)

| Feature Dimension | Type | Piiranha | Microsoft Presidio | Starpii |
|---|---|---|---|---|
| Basic Entity Recognition | Type 1 | 22% | 3% | 54% |
| Contextual Entity Disambiguation | Type 1 | 35% | 9% | 71% |
| Noisy and Real-World Data Detection | Type 1 | 7% | 1% | 26% |
| Evolving & Novel Entities Detection | Type 1 | 9% | 1% | 33% |
| Cross-Lingual or Multi-Lingual NER | Type 1 | 21% | 0.9% | 36% |
| *Adversarial | Type 1 | 19% | 4% | 69% |
| Basic Entity Recognition | Type 2 | 39% | 12% | 91% |
| Contextual Entity Disambiguation | Type 2 | 56% | 21% | 98% |

| Feature Dimension | Type | Piiranha | Microsoft Presidio | Starpii |
|---|---|---|---|---|
| Noisy and Real-World Data Detection | Type 2 | 12% | 3% | 56% |
| Evolving & Novel Entities Detection | Type 2 | 21% | 6% | 70% |
| Cross-Lingual or Multi-Lingual NER | Type 2 | 39% | 1% | 79% |
| *Adversarial | Type 2 | 24% | 7% | 81% |

Table 4 depicts the PII-Type-wise trends in non-identification of any PII in the input text. Notably, bank UPI Ids have been misclassified as email IDs etc., hence the low non-identification numbers. This demonstrates how such systems may fail to detect emerging and novel types of sensitive information. Our dataset consisted of diverse representations and syntax variations. The high non-identification values for credit card numbers, National Identity Numbers, Vehicle Registration Numbers and phone numbers may be attributed to the lack of diversity in the respective training datasets and outdated representations. Another reason for these high numbers may be the fact that our test dataset incorporates PII from several geographies across continents which are often undermined during data collection.

Table 4: **PII Type-wise trends of non-identification of any PII in the text by the models (represented as % of instances)**

| PII Type | Description | Piiranha | Microsoft Presidio | Starpii |
|---|---|---|---|---|
| Bank account | Bank account numbers | 30.0% | 5.1% | 76.5% |
| Bank UPI id | Bank digital transactional ID used in India | 4.8% | 1.0% | 26.3% |
| Credit card | Credit card numbers with digit variations | 34.3% | 18.1% | 64.0% |
| Date of Birth | Date of Birth in varied formats | 29.4% | 6.2% | 62.6% |
| Driver license | License number | 5.7% | 0.8% | 78.0% |
| Email | Email id in varied representations | 2.0% | 1.8% | 22.1% |
| Insurance Number | National Insurance Number | 0.0% | 0.0% | 85.9% |
| Name | Person names | 10.1% | 7.4% | 23.0% |
| Names of places or nouns | Books, mythological characters, movies etc | 18.0% | 5.9% | 26.2% |
| National Identity - SSN, Aadhar | Social Security Number, Aadhar, AHV Nummer, Número de Identificação da Segurança Social etc | 35.6% | 1.9% | 74.3% |
| Other National Identity | Residence ID, citizenship certificate, other national identities | 32.1% | 0.2% | 79.4% |
| Passport number | Passport number in varied countries | 47.4% | 1.0% | 77.7% |
| Phone | Phone number with and without the country code | 33.7% | 2.8% | 77.5% |
| Postal code | Zip code, postal code and PIN code | 0.7% | 1.7% | 84.1% |
| Tax identification | Tax identification number, PAN, Numer Identyfikacji Podatkowej, Número de Identificación Fiscal, etc | 28.8% | 1.5% | 77.0% |
| Vehicle registration | Vehicle registration number in different geographies | 61.4% | 5.7% | 88.2% |

## 3.2 Misclassification of PII in the text by the models

Table 5 shows the proportion of instances where the seed PII, although detected, was misclassified by the models. The models failed to correctly classify the PII in 67% of test cases. The misclassification rate is quite high across all the models, especially Presidio and Starpii. Notably, we do not find a significant difference in the misclassification rate between the two types of sentences (Type 1 has explicit reference to the type of PII while Type 2 does not). This observation signifies that the misclassification stems not just from context variation, possibly occurring due to lack of variation in the training data.

Table 5: Feature Dimension **wise trends of misclassification of seed PII sample in the text by the models (represented as % of instances)**

| Feature Dimension | Type | Piiranha | Microsoft Presidio | Starpii |
|---|---|---|---|---|
| Basic Entity Recognition | Type 1 | 51.6% | 82.8% | 84.9% |
| Contextual Entity Disambiguation | Type 1 | 37.2% | 92.3% | 88.4% |
| Noisy and Real-World Data Detection | Type 1 | 29.7% | 84.2% | 80.2% |
| Evolving & Novel Entities Detection | Type 1 | 51.9% | 83.6% | 87.7% |
| Cross-Lingual or Multi-Lingual NER | Type 1 | 41.5% | 82.0% | 81.7% |
| *Adversarial | Type 1 | 40.4% | 92.2% | 91.2% |
| Basic Entity Recognition | Type 2 | 41.8% | 85.9% | 84.4% |
| Contextual Entity Disambiguation | Type 2 | 100.0% | 90.2% | 93.2% |
| Noisy and Real-World Data Detection | Type 2 | 47.7% | 83.7% | 87.4% |
| Evolving & Novel Entities Detection | Type 2 | 31.3% | 87.4% | 86.3% |
| Cross-Lingual or Multi-Lingual NER | Type 2 | 39.7% | 91.6% | 91.2% |
| *Adversarial | Type 2 | 44.0% | 91.6% | 93.1% |

## 4 Discussion

### 4.1 Gaps in current PII masking models

The results of this research are meant to show a cumulative assessment of the models across various scenarios to present the need for a comprehensive evaluation framework. The analysis revealed that there were both non-identification cases and substantial misclassifications, especially for Presidio and Starpii. Further, on analyzing the misclassifications we noted that the models misclassified Bank accounts as Date, PAN, person, phone number, driver license, passport; Bank UPI ids as Date, email, PAN, location, person, Credit card as Date, location, person, Date of birth as Building, PAN, location, person, Bank account, driver license; Email as Date, location, person; and Name as Email, date, city, PAN, telephone, location. This also expresses that the models that utilize regular expressions and simplistic approaches may not be able to detect emerging variations of the PII (short codes, symbol representations etc.) [15]. However, models face challenges in nuanced language, or implicit references, leading to inaccurate PII identification. Furthermore, there exists complexity in distinguishing between names of people, organizations, and locations can be challenging, especially with common names or acronyms. Also, new slang, abbreviations, and online handles constantly emerge, requiring models to be continuously updated.

Traditional anonymization approaches treat all types of PII equally. However, the level of vulnerability of exposure for a Social Security Number and a Social Media User Id may not be the same. Hence, such a one-size-fits-all approach is not effective in protecting user privacy in a fine-grained manner [3]. Further, models built on large language models may inadvertently leak information or expose the type of training if they are overfitted during the fine-tuning process. Hence caution shall be taken to curate and synthesis the data prior to training or fine-tuning of the models [16]. In addition, research also suggests that many name detection algorithms exhibit biases that can lead to unequal protection across different demographic groups [17]. This is also attributable to lack of demographic diversity in the training data.

Consequent to the above discussion, despite masking efforts, there is potential for residual information that can be exploited to re-identify individuals, especially when combined with other data sources. This risk underscores the need for robust anonymization techniques and continuous evaluation of masking effectiveness [16]. Further, there is a lack of consistent metrics for evaluating PII masking and the publicly available datasets that could be used for benchmarks are limited in scope and coverage to truly evaluate PII masking.

### 4.2 Implications of privacy from above failures

PII masking approaches based on transformer-based models, though adept at capturing long-range dependencies, often falter in accurate entity recognition due to their struggle with local feature extraction. Local features, such as capitalization, punctuation, prefixes, suffixes, and surrounding word sequences, are vital for discerning entities, especially in the context of personally identifiable information (PII). In the context of PII, these features contain specific contextual information crucial for accurate masking. However, during the preprocessing stage of LLM-based scraping approaches, mark-down, stemming, and tokenization can lead to the loss of these crucial identifiers, thereby limiting the possibility of PII detection and masking. Furthermore, the claims by large language model developers that they had removed PII prior to training are questionable, as the effectiveness of such pre-training removals is unclear.

Given that the PII masking is a vital component of data privacy strategies in today's digital landscape (for enabling data access, training models, consumer communications and agentic systems), failure to appropriately detect the PII and subsequent exposure of such PII may lead to financial loss, reputational loss and in some cases loss of emotional identity for corporations managing the PII and the individuals whose PII is in question.

### 4.3 Need for accountability and transparency

While evaluating Personally Identifiable Information (PII) masking models, the choice of metrics can significantly impact the perceived limitations of Named Entity Recognition (NER) systems. While precision and recall are commonly used to assess NER systems, these metrics alone may not capture the full picture of a model's effectiveness in practical applications. [18] mentions that a narrow focus on these metrics can lead to an inaccurate understanding of a system's performance in real-world scenarios, where the context and relevance of recognized entities are crucial. Therefore, when evaluating PII masking models, it is essential to consider metrics that reflect the broader context and utility of entity recognition in downstream tasks to gain a more comprehensive understanding of the models' effectiveness and limitations therein.

In addition, all the three models stated that they were evaluated based on accuracy, precision, recall and F1 scores. None of the models detailed any limitations of using this model. All the models provided a disclaimer that their model results may be incorrect, and they assume no responsibility for such incorrect outcomes. Given the above context, the models built on PII masking need to have sufficient disclosure on the extent of tests that were carried out on these models, the validation results and the limitations of the model, to enable the downstream uses of the model to be well informed prior to adoption of the model for PII masking tasks.

Hence, there is a need for increased responsibility for corporations relying on and / or researchers building datasets and models need to be more sensitive towards the downside impacts of the exposure caused by PII masking failures.

## 5 Way Forward

### 5.1 Limitations

We have four broad limitations to our research. They are: (a) The curated variations of PII are not exhaustive. For instance, we have not considered abbreviations and PII within code; (b) While we made attempts to ensure that the seed PII samples created using language models are free from errors, there may be errors in such generation; (c) The text generated using the language models are not consistent, there may be errors in the text. We only validated if the text contained the seed PII sample or otherwise using regex-based methods. We did not verify the appropriateness of the text generated, beyond sample validation; and (d) We adopted text generation by instructing the language models (using text prompts) to generate text that masks or does not disclose the PII type. There may be errors in these generations and the model may not have masked the type of PII in text generated for 'Type 2' feature dimensions.

## 5.2 Way Forward

As a way forward we have 3 key suggestions for researchers looking to contribute to this area of research. Firstly, we recommend expanding dataset diversity by including more jurisdictions and PII types in the dataset, compiling dataset for PII within code (email id split using javascript) and building benchmarks for evaluating the PII masking models. Secondly, we believe better adversarial text generation for PII can be evolved and tested on PII masking models. Future research may consider the development of adversarial testing frameworks for PII masking. And finally, we recommend inter-disciplinary research for addressing performance gaps and extending model capabilities for PII masking models.

To enable better research, we are open sourcing the availability of the curated dataset for the research community.

# REFERENCES

[1] Badgujar, P. (2025). Implementing Data Masking Techniques for Privacy Protection. Journal of Technological Innovations, 2(4). https://doi.org/10.93153/5yysvh44

[2] C., K., & S., S. (2016). Privacy preserved data publishing techniques for tabular data. International Journal of Computer Applications, 151(1), 1–6. https://doi.org/10.5120/ijca2016911874

[3] Majeed, A., Ullah, F., & Lee, S. (2017). Vulnerability- and diversity-aware anonymization of personally identifiable information for improving user privacy and utility of publishing data. Sensors, 17(5). https://doi.org/10.3390/s17051059

[4] Ehrmann, M., Hamdi, A., Pontes, E. L., Romanello, M., & Doucet, A. (2023). Named entity recognition and classification in historical documents: A survey. ACM Computing Surveys, 56. https://doi.org/10.1145/3604931

[5] Zhang, Z. (2013). Named entity recognition: Challenges in document annotation, gazetteer construction, and disambiguation. Semantic Scholar. https://api.semanticscholar.org/CorpusID:33743976

[6] Beltagy, I., Lo, K., & Cohan, A. (2019). SCIBERT: A pretrained language model for scientific text. Proceedings of the 2019 Conference on Empirical Methods in Natural Language Processing and 9th International Joint Conference on Natural Language Processing, 3615–3620. https://doi.org/10.18653/v1/d19-1371

[7] N., D., & Bhadka, H. (2017). A survey on various approaches used in named entity recognition for Indian languages. International Journal of Computer Applications, 167(11), 11–18. https://doi.org/10.5120/ijca2017913878

[8] Stanislawek, T., Wróblewska, A., Wójcik, A., Ziembicki, D., & Biecek, P. (2019). Named entity recognition: Is there a glass ceiling? Proceedings of the 23rd Conference on Computational Natural Language Learning (CoNLL 2019), 624–633. https://doi.org/10.18653/v1/k19-1058

[9] Fadnavis, R.A. (2020). A Review: Approaches and Challenges for Named Entity Recognition for Indian Language Text. Journal of emerging technologies and innovative research. https://api.semanticscholar.org/CorpusID:215797830

[10] AI4Privacy. (2023). pii-masking-200k (Revision 1d4c0a1) [Dataset]. Hugging Face. https://doi.org/10.57967/hf/1532

[11] bigcode. (n.d.). bigcode-pii-dataset [Dataset]. Hugging Face. https://huggingface.co/datasets/bigcode/bigcode-pii-dataset

[12] iiiorg (n.d.). piiranha-v1-detect-personal-information [Dataset]. Hugging Face. https://huggingface.co/iiiorg/piiranha-v1-detect-personal-information

[13] bigcode. (n.d.). starpii [Dataset]. Hugging Face. https://huggingface.co/bigcode/starpii

[14] Mendels, O., Peled, C., Vaisman Levy, N., Hart, S., Rosenthal, T., Lahiani, L., & others. (2018). *Microsoft Presidio: Context-aware, pluggable, and customizable PII anonymization service for text and images*. Microsoft. https://microsoft.github.io/presidio

[15] Liu, Y., Liao, L., & Song, T. (2020). Static tainting extraction approach based on information flow graph for personally identifiable information. Science China Information Sciences, 63. https://doi.org/10.1007/s11432-018-9839-6

[16] Lukas, N., Salem, A., Sim, R., Tople, S., Wutschitz, L., & Zanella-Béguelin, S. (2023). Analyzing leakage of personally identifiable information in language models. Proceedings of the IEEE Symposium on Security and Privacy, 2023, 346–363. https://doi.org/10.1109/SP46215.2023.10179300

[17] Mansfield, C., Paullada, A., & Howell, K. (2022). Behind the mask: Demographic bias in name detection for PII masking. Proceedings of the 2nd Workshop on Language Technology for Equality, Diversity and Inclusion (LTEDI 2022), 76–89. https://doi.org/10.18653/v1/2022.ltedi-1.10

[18] Kocaman, V., & Talby, D. (2022). Accurate clinical and biomedical named entity recognition at scale. *Software Impacts, 13*, 100373. https://doi.org/10.1016/j.simpa.2022.100373